\documentclass{article}

\PassOptionsToPackage{numbers, compress}{natbib}
\usepackage[preprint]{neurips_2024}
\usepackage{graphicx,amsmath}
\usepackage[utf8]{inputenc} 
\usepackage[T1]{fontenc}    
\usepackage[hidelinks]{hyperref}       
\usepackage{url}            
\usepackage{booktabs}       
\usepackage{amsfonts}       
\usepackage{nicefrac}       
\usepackage{microtype}      
\usepackage{xcolor}         
\usepackage{multirow}
\usepackage{pifont}
\usepackage{multirow}
\usepackage[font=small,labelfont=bf]{caption}

\hypersetup{
    colorlinks=true,
    linkcolor=blue,
    filecolor=blue,      
    urlcolor=blue,
    citecolor=blue,
}

\title{CrossCheckGPT: Universal Hallucination\\ Ranking for Multimodal Foundation Models}

\author{%
Guangzhi Sun$^{1*}$ \quad Potsawee Manakul$^{1,2,3*}$ \quad Adian Liusie$^1$ \quad Kunat Pipatanakul$^{2,3}$ \\  \textbf{Chao Zhang$^4$} \quad \textbf{Phil Woodland$^1$} \quad \textbf{Mark Gales$^1$} \vspace{2.5mm} \\ 
\vspace{2.5mm}
$^1$University of Cambridge \quad $^2$SCB 10X \quad $^3$SCBX  \quad $^4$Tsinghua University \\ 
\texttt{gs534@cam.ac.uk}, \texttt{potsawee@scb10x.com}, \texttt{al826@cam.ac.uk}
}

\begin{document}

\maketitle
\def\thefootnote{*}\footnotetext{Equal contribution}\def\thefootnote{\arabic{footnote}}

\vspace{-2mm}
\begin{abstract}
Multimodal foundation models are prone to hallucination, generating outputs that either contradict the input or are not grounded by factual information. Given the diversity in architectures, training data and instruction tuning techniques, there can be large variations in systems’ susceptibility to hallucinations. To assess system hallucination robustness, hallucination ranking approaches have been developed for specific tasks such as image captioning, question answering, summarization, or biography generation. However, these approaches typically compare model outputs to gold-standard references or labels, limiting hallucination benchmarking for new domains. This work proposes "CrossCheckGPT", a reference-free universal hallucination ranking for multimodal foundation models. The core idea of CrossCheckGPT is that the same hallucinated content is unlikely to be generated by different independent systems, hence cross-system consistency can provide meaningful and accurate hallucination assessment scores. CrossCheckGPT can be applied to any model or task, provided that the information consistency between outputs can be measured through an appropriate distance metric. Focusing on multimodal large language models that generate text, we explore two information consistency measures: CrossCheck-explicit and CrossCheck-implicit. We showcase the applicability of our method for hallucination ranking across various modalities, namely the text, image, and audio-visual domains. Further, we propose the first audio-visual hallucination benchmark, "AVHalluBench", and illustrate the effectiveness of CrossCheckGPT, achieving correlations of 98\% and 89\% with human judgements on MHaluBench and AVHalluBench, respectively. 

\end{abstract}


%


\section{Introduction}
\label{sec:intro}

In the domain of generative foundation models, ‘hallucination’ describes the scenario when generated outputs, while seemingly credible, are either inconsistent with the provided context or contradict established factual knowledge \cite{avhallucination_1,avhallucination_2,avhallucination_3}. This issue impacts many generative applications and can lead to the spread of misinformation in a range of settings \cite{hallucination_risk,hallucination_risk_2}. Given the differences in architectures, data, and alignment techniques for foundation models, there is a need to be able to quantify a system’s susceptibility to hallucination, such that practitioners can be aware of systems’ hallucination risk and select systems with high factual consistency.

\begin{figure}[t]
    \centering
    \includegraphics[scale=0.5]{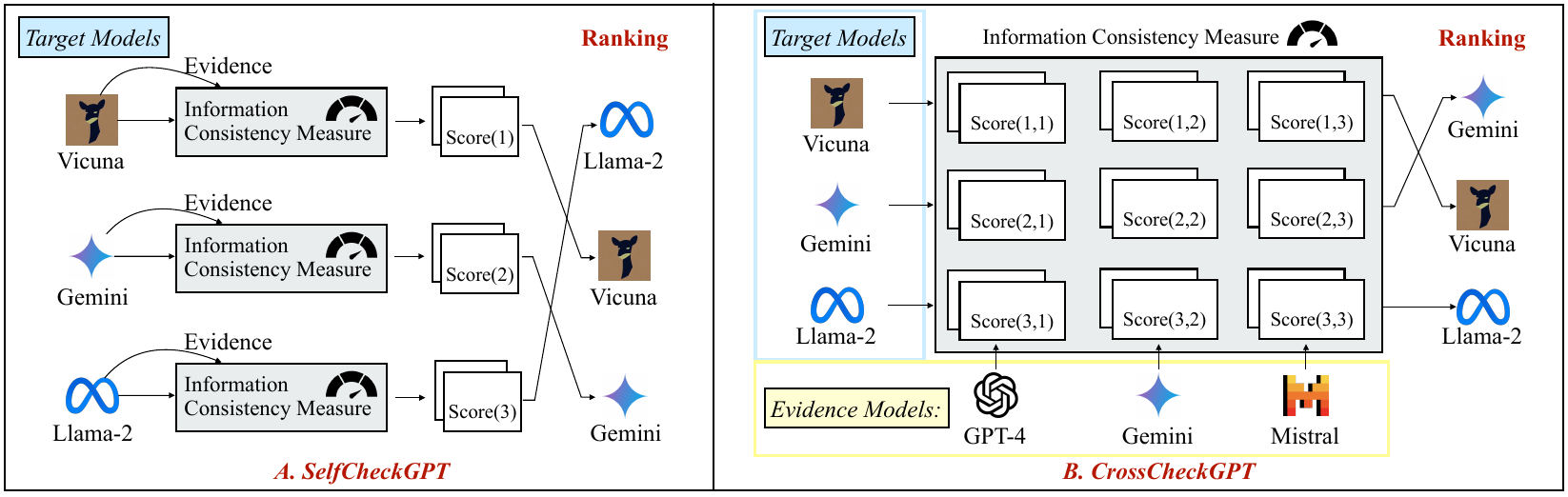}
    \caption{SelfCheckGPT (Left) and CrossCheckGPT (Right) for hallucination rankings. The approach can rank a set of MLLMs on any task without reference, enabling hallucination benchmarks for various generative tasks.}
    \vspace{-0.3cm}
    \label{fig:overview}
\end{figure}

Current hallucination benchmarks have been developed to rank systems for individual tasks including question answering \cite{truthfulqa, triviaqa, halueval, faithdial,thorne-etal-2018-fever}, summarization \cite{maynez-etal-2020-faithfulness, manakul-etal-2023-mqag}, biography generation \cite{manakul2023selfcheckgpt}, instruction following \cite{memotrap}, image captioning \cite{chair}, and visual question answering \cite{pope,amber}. Many of these benchmarks measure the hallucination level through a proxy measure, such as the ability of the model to correctly answer questions designed to trigger hallucinations. However, these benchmarks have been designed for particular tasks and assume access to gold-standard labels, limiting their applicability to generalized domains. On the other hand, hallucination detection approaches such as SelfCheckGPT \cite{manakul2023selfcheckgpt} and UniHD \cite{chen2024unified} directly examine generated responses against self-evidence, and therefore do not require gold-standard answers. These methods, though, simply aim to identify when a model hallucinates, and scores are not directly comparable across different models. 

In this paper, we propose CrossCheckGPT, a universal hallucination ranking approach to benchmark multimodal foundation models.
The core idea of CrossCheckGPT is that the same hallucinated content is unlikely to be generated by different independent systems, while factual content likely to be consistent across models.
An illustration of the approach and its contrast to SelfCheckGPT is depicted in Fig.~\ref{fig:overview}. Instead of checking for self-consistency, as done in SelfCheckGPT, CrossCheckGPT checks the \textit{cross-consistency} by comparing against evidence generated from a set of independent models. This produces more accurate and directly comparable hallucination scores, as well as yielding more robust rankings. 
CrossCheckGPT can be applied to any foundation model and task as long as a suitable information consistency measure is used. 
%
%
This paper demonstrates the effectiveness of CrossCheckGPT as a \textit{universal} evaluation framework for any Multimodal Large Language Model (MLLM) that generates text outputs, applicable irrespective of the input modality. We investigate two information consistency measures: CrossCheck-explicit, which generates multiple text samples from each evidence system, and CrossCheck-implicit, which prompts the evidence model to determine whether it agrees with the assessed outputs.    

CrossCheckGPT is validated on WikiBio \cite{manakul2023selfcheckgpt} and MHaluBench \cite{chen2024unified} as text-to-text and image-to-text description tasks, and our experiments show that CrossCheckGPT achieves a notable 98\% Spearman's Rank Correlation (SRC) on MHaluBench against human ranking compared to -10\% SRC using SelfCheckGPT and 33\% using UniHD. In addition, a comprehensive audio-visual hallucination benchmark dataset (AVHalluBench) is proposed, covering a diverse range of styles, domains and elements such as visual text, speech and music. The AVHalluBench is used to rank recent audio and video LLMs such as Gemini 1.5 Pro, conducting the first study on audio-visual hallucination benchmarking. The key contributions of this paper are summarized as follows:

\begin{itemize}
    \item We propose CrossCheckGPT, a reference-free hallucination ranking approach that can be applied universally across text-generation tasks for systems of different modalities. 

    \item We conduct comprehensive experiments over a range of tasks and modalities, demonstrating the effectiveness of CrossCheckGPT as a hallucination benchmarking approach for ranking text, image or audio-visual systems. Experimental results illustrate that CrossCheckGPT consistently outperforms alternate approaches, such as SelfCheckGPT \cite{manakul2023selfcheckgpt} and UniHD \cite{chen2024unified}.
    \item We analyze hallucination within video understanding and curate AVHalluBench, which to the best of our knowledge, is the first publicly released audio-visual hallucination benchmark.
\end{itemize}

\section{Related Work}
\textbf{LLM Hallucination Benchmarking}: Hallucination benchmarks typically rely on proxy tasks to probe the likelihood of LLM making factual errors. For example, question-answering (QA) based benchmarks, such as TriviaQA \cite{triviaqa}, TruthfulQA \cite{truthfulqa}, HaluEval-QA \cite{halueval}, MemoTrap \cite{memotrap} and FEWL \cite{similar_work} design questions specifically to probe truthfulness and factual accuracy and rank systems by their accuracy. 
Other methods, such as FaithDial \cite{faithdial}, XSum \cite{xsum} and CNN-DM \cite{cnndm} measure hallucination in dialogue responses or summarization.
However, these benchmarks require references (e.g., ground-truth answers or gold-standard references) to compare to model-generated outputs. On the other hand, SelfCheckGPT \cite{manakul2023selfcheckgpt} can be used to rank systems on hallucination levels by measuring systems' self-consistency scores on equivalent tasks. However, SelfCheckGPT was designed as a hallucination detection method and may not be calibrated across systems.

\textbf{Multimodal LLM Hallucination Benchmarking}: 
Multimodal hallucination has been mainly explored in the image-to-text domain for visual LLMs. One stream of methods, including CHAIR~\cite{chair}, LURE~\cite{zhou2024analyzing} and {MHaluBench}~\cite{chen2024unified}, directly evaluate the generated text descriptions of images using gold-standard annotations or external toolkits. Another stream of methods, such as {POPE} \cite{pope} and {HallusionBench}~\cite{hallusionbench}, curate a set of questions with short answers trying to capture various aspects of hallucination. Meanwhile, AMBER~\cite{amber} combines both generation and question answering in one single benchmark. Unlike these methods, CrossCheckGPT does not rely on gold-standard reference or dedicated question sets, and can be universally applied to any input modalities.
\section{CrossCheckGPT}
\label{sec:method}

\begin{figure}[h]
    \centering
    \includegraphics[scale=0.39]{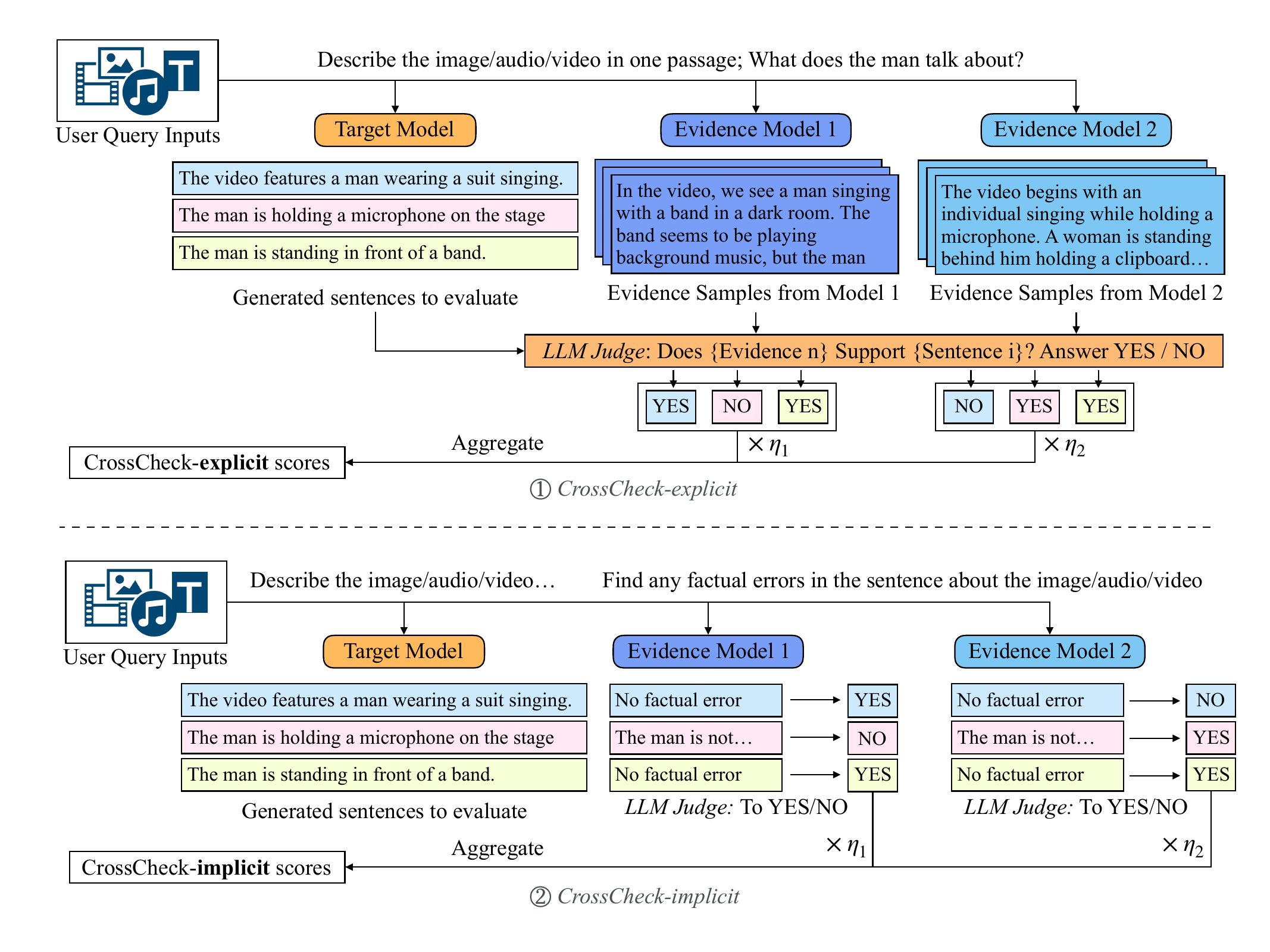}
    \caption{Illustration of the CrossCheckGPT approach with two evidence models as an example. Two information consistency measures are shown. \textcircled{\raisebox{-1.2pt}{1}} CrossCheck-explicit where $N$ passages are stochastically generated by sampling from each evidence model and \textcircled{\raisebox{-1.0pt}{2}} CrossCheck-implicit where evidence models are directly used to determine whether there are any factual errors in each sentence (without sampling). The LLM judge uses the sentence and the analysis from the evidence model to produce the Yes/No binary decision.}
    \vspace{-0.3cm}
    \label{fig:crosscheck}
\end{figure}

CrossCheckGPT assigns a score to an MLLM (denoted as the \emph{target} model) by assessing how much the responses of the MLLM are supported by evidence generated from a set of MLLMs (denoted as \emph{evidence models}). The CrossCheckGPT scores can then be used to rank the MLLMs. As illustrated in Fig.~\ref{fig:crosscheck}, we explore two information consistency measures, CrossCheck-explicit and CrossCheck-implicit, which measure the hallucination of generated responses either through the explicit generation of evidence passages or implicit prompting, respectively. CrossCheckGPT is reference-free and can be generally applied to MLLMs of any input modality and output response type.
\subsection{Information Consistency Measures}
\label{sec:measures}
CrossCheck-explicit stochastically generates a set of evidence passages from each evidence model and computes the average distance between each evidence passage and the target response. Let $R=[r_1,\ldots, r_i,\ldots, r_I]$ denote the response of the target model $\hat{M}$, where $r_i$ is the $i$-th sentence of the response, to a given query $Q$, which can be of any modality. We first re-formulate the SelfCheckGPT score for sentence $r_i$ of the target model in Eqn. \eqref{eq:selfcheck} below,
\begin{align}
    \mathcal{S}_\text{selfcheck}(\hat{M})&=\frac{1}{|\mathcal{Q}|}\frac{1}{I} \sum_{Q\in |\mathcal{Q}|} \sum_{i=1}^I \mathcal{S}^\text{selfcheck}_{r_i,Q}(\hat{M})\qquad \text{~where~} \; \mathcal{S}^\text{selfcheck}_{r_i,Q}(\hat{M})=\frac{1}{\hat{N}}\sum_{n=1}^{\hat{N}}x^{(n)}_{r_i,Q}(\hat{M})
    \label{eq:selfcheck}
\end{align}
where $\mathcal{Q}$ is the set of queries in a test set, $\hat{N}$ is the number of stochastically generated passages by the model $\hat{{M}}$, and $x^{(n)}_{r_i,Q}(\hat{{M}})$ denotes the hallucination score of whether sentence $r_i$ is supported by evidence $n$ from $\hat{{M}}$. The hallucination score, estimated by prompting an LLM judge with the sentence and each evidence, takes a value in $\{0, 1\}$, where $0$ denotes \textit{supported} and $1$ denotes \textit{hallucinatory}. 

\textbf{CrossCheck-explicit}, in contrast to SelfCheckGPT, uses the evidence from $|\mathcal{M}|$ evidence models and measures the distance of the response against those from all other systems. The overall CrossCheck-explicit score $\mathcal{C}_\text{explicit}(\hat{M})$ for a specific target model $\hat{M}$ can be computed using Eqn.~\eqref{eq:crosscheck},
\begin{equation}
    \mathcal{C}_\text{explicit}(\hat{M})\!=\!\frac{1}{|\mathcal{Q}|}\frac{1}{I} \!\! \sum_{Q\in |\mathcal{Q}|} \sum_{i=1}^{I} \mathcal{C}^\text{explicit}_{r_i,Q}(\hat{M}) \;\;\; \text{~where~} \; \mathcal{C}^\text{explicit}_{r_i,Q}(\hat{M}) \!=\!\frac{\sum_{j=1}^{|\mathcal{M}|}\eta_{j} \sum_{n=1}^{N_j} x^{(n)}_{r_i,Q}({M}_j)}{\sum_{j=1}^{|\mathcal{M}|}\eta_jN_j}
    \label{eq:crosscheck}
\end{equation}
where $\mathcal{M}$ denotes the set of evidence models used for CrossCheck-explicit. Note that self-consistency can be taken into account by including the target model $\hat{M}$ into the evidence models, $\hat{M}\!\in\! \mathcal{M}$. Each evidence model ${M}_j$ stochastically generates $N_j$ passages to check the response against, and since systems may have different levels of reliability, a factor $\eta_j$ can be assigned to the passages generated from model ${M}_j$. 
\textbf{CrossCheck-implicit} is an alternative consistency measure, where instead of explicitly generating passages for the same query, the evidence models are prompted to spot any factual errors in each sentence. The overall implicit CrossCheck-implicit score is computed using Eqn. \eqref{eq:crosscheckimp},
\begin{equation}
\mathcal{C}_\text{implicit}(\hat{M})=\frac{1}{|\mathcal{Q}|}\frac{1}{I}\sum_{Q\in |\mathcal{Q}|} \sum_{i=1}^{I}\mathcal{C}^\text{implicit}_{r_i,Q}(\hat{M})\qquad\text{~where~}\;\;\mathcal{C}^\text{implicit}_{r_i,Q}(\hat{M})=\sum_{j=1}^{|\mathcal{M}|}\eta_{j}\,y_{r_i,Q}({M}_j)
    \label{eq:crosscheckimp}
\end{equation}
where $y_{r_i,Q}({M}_j)$ denotes the hallucination score of sentence $r_i$ computed using CrossCheck-implicit. In contrast to CrossCheck-explicit (which computes $x_{r_i,Q}({M}_j)$), $y_{r_i,Q}({M}_j)$ is computed by first prompting the evidence model $M_j$ to analyze whether $r_i$ contains any factual errors given the input $Q$. The LLM judge then takes the input $r_i$ and analysis from model $M_j$ and predicts $y_{r_i,Q}({M}_j)$, whether the response is hallucinatory. If factual errors are found in $r_i$, $y_{r_i,Q}({M}_j)=1$, and otherwise $y_{r_i,Q}({M}_j)=0$. 
%
We note that concurrent work, PoLL \cite{verga2024replacing}, applies a group of models as judges to evaluate texts and can be viewed as similar to CrossCheck-implicit. This work focuses on multimodal inputs and hallucination benchmarking.

\subsection{Confidence-based Weighting for Evidence Models}

While all evidence models are advanced MLLMs, the quality of their evidence may vary depending on their propensity to hallucinate. Therefore, a weighting mechanism is proposed where the scores are weighted by model uncertainty reflected by SelfCheckGPT scores, as shown below:
\begin{equation}
    \eta_j = \frac{e^{-\mathcal{S}_\text{selfcheck}({M}_j)/T}}{\sum_{k=1}^{|\mathcal{M}|} e^{-\mathcal{S}_\text{selfcheck}({M}_k)/T}},
    \label{eq:weighting}
\end{equation}
where $T$ is the calibration temperature that determines the sharpness of the weight distribution, which is set to a constant for each benchmark. 
A higher SelfCheckGPT score indicates that the model tends to generate inconsistent information and is more uncertain. In addition, this weighting mechanism ensures that outlier systems will not be undermined by the evidence from weaker models.\footnote{Note that a weight distribution can also be associated with each specific query by using the average SelfCheckGPT score of each evidence model.}

\section{CrossCheckGPT for Hallucination with Multimodal Inputs}
CrossCheckGPT is designed to be general and applicable to models of any input modality, provided that the outputs are of a consistent form (i.e. text) and a suitable information consistency measure is used. This general design of CrossCheckGPT enables it to also be applied to rank multi-modal systems (i.e. systems which use two or more input modalities). 

\begin{figure}[h]
    \centering
    \includegraphics[scale=0.42]{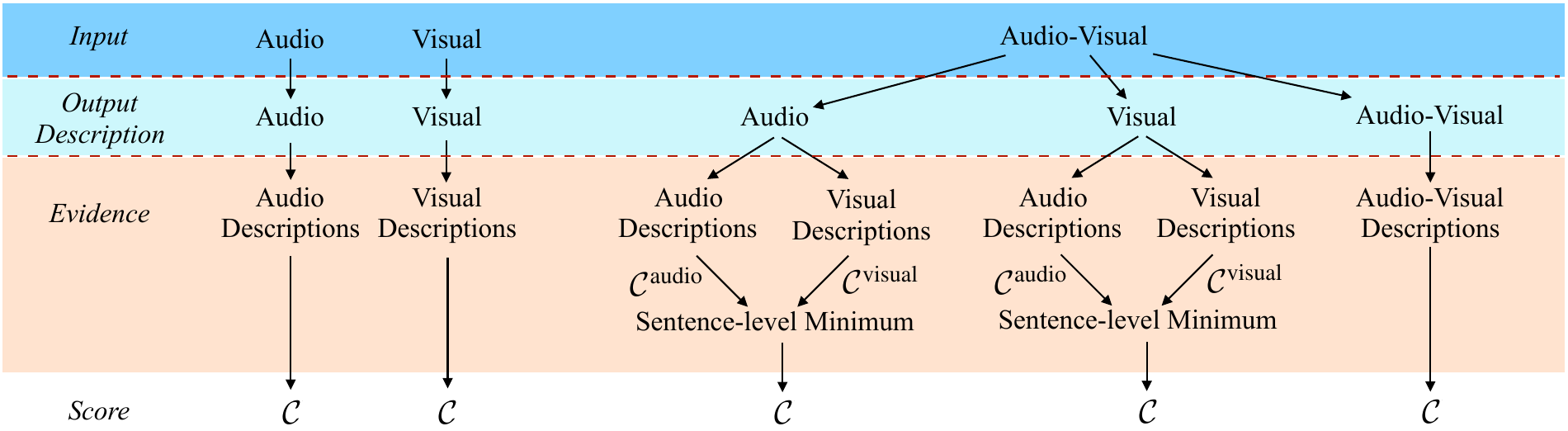}
    \caption{CrossCheckGPT score computation for AVHalluBench with audio, visual and audio-visual inputs.}
    \vspace{-0.2cm}
    \label{fig:avscore}
\end{figure}

As shown in Fig. \ref{fig:avscore}, we use CrossCheckGPT to evaluate models of three different categories: the \emph{audio} domain and \emph{visual} domain where the inputs are either audio or visual (image or silent video), and we further conduct the first study on evaluating hallucination levels within the \emph{audio-visual} domain where the inputs are videos with their paired audio. Due to the lack of diversity in current publicly available capable systems taking audio-visual inputs, to evaluate CrossCheckGPT in the audio-visual domain, we prompt multi-modal models to instead 
split the outputs into visual descriptions and auditory descriptions, evaluating CrossCheckGPT within either of the domains. We use visual descriptions to check the visual-only inputs and audio descriptions to check the audio-only inputs. For hallucination benchmarking in multimodal audio-visual settings, information may require both modalities, e.g. someone demonstrating and explaining a skateboard trick. 
In this scenario, we use $\mathcal{C}= \min \left(\mathcal{C}^\text{audio}, \mathcal{C}^\text{visual}\right)$ as the CrossCheckGPT score,
where $\mathcal{C}^\text{audio}$ uses the audio descriptions and $\mathcal{C}^\text{visual}$ uses the visual descriptions.\footnote{For simplicity, $\hat{M}$, $r_i$, and $Q$ are dropped here, and the scores can be either implicit or explicit.}\footnote{Initial findings showed CrossCheck-implicit gives different ranges of scores for audio and visual modalities, at about 0.2 and 0.5 on average, respectively. Thus, only CrossCheck-explicit is adopted for audio-visual inputs.}

\textbf{AVHalluBench}: To benchmark hallucinations in audio-visual LLMs, we curate AVHalluBench, a dataset containing 175 videos selected from six video understanding datasets covering various styles and elements, with statistics shown in Table~\ref{tab:AVHalluBench1} in the Appendix. To verify the effectiveness of CrossCheckGPT (and future benchmarking methods), AVHalluBench includes a carefully written set of hallucination-free descriptions for audio and visual contents. After watching each video with audio, the annotators were instructed to write \textit{one} description focusing on the audio content and \textit{one} description focusing on the visual content of the video, \textit{separately}.\footnote{To maximize coverage, initial descriptions were generated using Gemini 1.5 Pro and GPT-4v, prompted to describe all the elements present in the sequence of frames. Note that although these descriptions are \textit{not} hallucination-free, they have a high level of coverage and subjective details. The annotators were provided with these descriptions in addition to the videos while being instructed to write only objective details of the videos.} To analyze the inter-annotator agreement, we split each description into atomic facts \cite{min-etal-2023-factscore} and verify each fact against the descriptions written by the other annotators, categorized as either: \textit{Supporting}, such that the fact is supported by the other annotator, \textit{Contradicting}, such that the fact contradicts the information provided by the other annotator, or \textit{Neutral} such that the facts neither support nor contradict one another. Both decomposition and verification processes are performed automatically using GPT-4. Of the 39 videos annotated by multiple annotators, there were 471 audio-related facts and 913 visual-related facts, and the agreement between annotators (as counted by Supporting/Neutral/Contradicting) was 64.6\%/24.6\%/10.8\% and 62.0\%/29.0\%/9.0\%, respectively.

\section{Experiments}
\label{sec:exp}

We conduct experiments to validate CrossCheckGPT on MLLMs with three input modalities, including text (\S\ref{section:text_only}), image (\S\ref{section:image_only}), and audio-visual (\S\ref{section:audiovisual}). During inference, we use a temperature of 1.0, a beam size of 1 and a top-p of 0.9 are used for all models. \textit{SelfCheckGPT}~\cite{manakul2023selfcheckgpt} is applied as a hallucination ranking baseline for all modalities since it is reference-free and not task-specific.



\subsection{Text-to-text Experiments}
\label{section:text_only}



\textbf{Experimental Setup}: The main text-to-text experiments are performed using the subset of WikiBio data used in \citep{manakul2023selfcheckgpt}, which contains 238 biographical passages from Wikipedia. We select 10 open-source LLMs (listed in Appendix Table \ref{tab:models}) as target models, 8 of which are used as evidence models. Four models are Llama-2-7B based \citep{touvron2023llama} (e.g. Vicuna-v1.5-7B \cite{vicuna}) and four models are Mistral-7B based \citep{jiang2023mistral}. Each evidence model generates 20 stochastic passages. For the LLM judge in CrossCheck-explicit (used to determine whether sentences support one another), Mistral-7B \cite{jiang2023mistral} is used as it achieves the best results among all considered open-source LLMs (shown in Appendix Table~\ref{tab:aucpr}).

To evaluate the \textit{general} benchmarking ability of ranking methods, 10 benchmark metrics from the hallucinations leaderboard\footnote{\url{https://huggingface.co/spaces/hallucinations-leaderboard/leaderboard}} (shown in Table \ref{tab:benchmark}) are selected to provide the overall hallucination ranking of the systems. These metrics are either based on human annotation or gold-standard references, where the overall rankings are obtained by averaging the rankings from each metric. 

We report the \emph{system-level} correlation between the hallucination ranking methods and the overall ranking measured by Spearman's Rank Correlation coefficient (SRC), denoted as System($\rho$). In addition, as WikiBio contains reference texts, the references can be used as evidence texts, which can be considered an idealized fact-checking method. This method is referred to as RefCheck, and CrossCheckGPT and SelfCheckGPT scores also are compared against RefCheck at \emph{document-level} using Pearson's Correlation Coefficient (PCC), denoted as Document$(r)$. Furthermore, to investigate the effectiveness of CrossCheckGPT when the target LLM is much more powerful than those evidence models, we include GPT-4 in addition to the 10 target LLMs.

\textbf{Hallucination Ranking Results}: Existing hallucination metrics such as HaluEval-QA accuracy do not correlate well with the overall ranking at the system level. Some metrics have negative correlations while the highest (TruthfulQA MC2) is 57.14\% (shown in Table~\ref{tab:weightcross}, with further pairwise correlations provided in Appendix Table~\ref{tab:metricssystem}). This is likely because each existing metric is typically designed to measure only one aspect related to hallucinations, e.g., probing through question-answering.

\begin{table}[h]
    \centering
    \begin{tabular}{p{5cm}ccc}
    \toprule
    \multirow{2}{*}{Metrics}     & \multirow{2}{*}{System($\rho$) (\%)} & \multicolumn{2}{c}{Document ($r$) (\%)} \\
                                 & &w/o GPT4 & with GPT4 \\ 
    \midrule

    TruthfulQA MC2 \cite{truthfulqa}      & 57.14  & - & - \\
    SelfCheckGPT \cite{manakul2023selfcheckgpt} & 66.46 & 74.06 & 76.08 \\
    \midrule
    CrossCheck-implicit & 56.71 & 18.33 & 17.29 \\
    CrossCheck-explicit & \underline{77.44} & \textbf{82.28} & \underline{77.23} \\
    CrossCheck-implicit weighted & 56.81 & 20.21 & 19.16 \\
    CrossCheck-explicit weighted & \textbf{82.32} & \underline{81.78} & \textbf{82.18} \\
    \bottomrule
    \end{tabular}
    \vspace{0.2cm}
    \caption{General hallucination evaluation where the task for SelfCheckGPT/CrossCheckGPT is open-ended biography generation on WikiBio. System-level correlation, System($\rho$), is measured against the overall ranking of the leaderboard, and document-level correlation, Document($r$), is measured against RefCheck. ``With GPT-4'' refers to including GPT-4 as a target model. Additional metrics are presented in Table \ref{tab:weightcross2} in the Appendix.}
    \vspace{-0.3cm}
    \label{tab:weightcross}
\end{table}

\begin{minipage}{\textwidth}
  \begin{minipage}[b]{0.67\textwidth}
    \centering
    \includegraphics[scale=0.52]{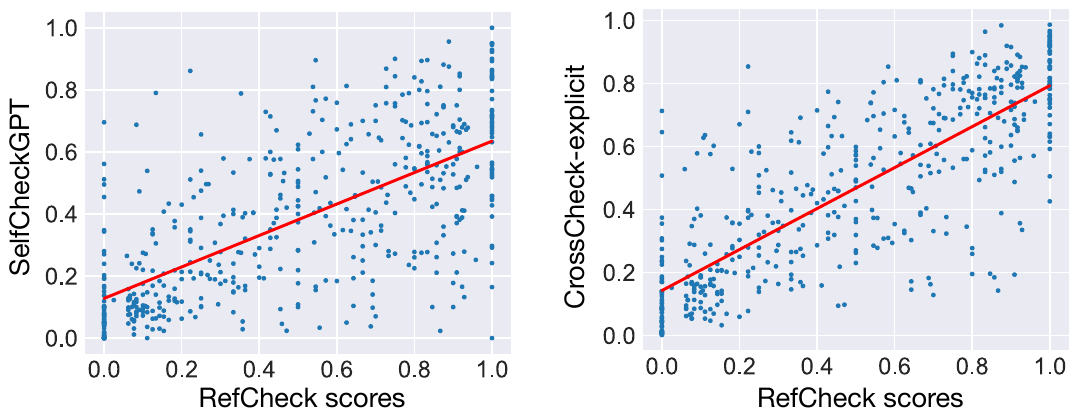}
    \captionof{figure}{Scatter plot of document-level scores for SelfCheckGPT and CrossCheck-explicit against RefCheck for text-to-text experiments.}
    \label{fig:textscatter}
  \end{minipage}
  \hfill
  \begin{minipage}[b]{0.3\textwidth}
    \centering
    \begin{tabular}{cc}
    \toprule
      Subset & Values \\
      \midrule
      Succ. Rate & 90\% \\
      P-value & 4$\times 10^{-6}$\\
        \bottomrule
      \end{tabular}
      \captionof{table}{Success rate of CrossCheck outperforming SelfCheck for independent subsets of WikiBio documents. The P-value is measured by the one-tailed sign test with $H_0=$ CrossCheck not better than SelfCheck.}
      \label{tab:textsignificance}
    \end{minipage}
  \end{minipage}

CrossCheck-explicit correlates with the overall ranking better than all other methods, with CrossCheck-explicit weighted by model uncertainty achieving the highest correlation, highlighting its effective \textit{general} hallucination ranking ability. 
In addition, the document-level correlation plots are shown in Fig. \ref{fig:textscatter}, and the sign test on independent subsets in Table \ref{tab:textsignificance} shows the statistical significance ($p=4\times 10^-6$) of CrossCheckGPT being better than SelfCheckGPT for ranking at the system-level.

\subsection{Image-to-text Experiments}
\label{section:image_only}
We validate CrossCheckGPT for the hallucination ranking of visual LLMs on image-to-text tasks. The experiments are performed on MHaluBench~\citep{chen2024unified}, an image-captioning hallucination dataset. Nine visual LLMs are selected as target models, all of which are used to generate evidence passages (see Appendix Table \ref{tab:models} for the list of models). Each evidence model generates ten image descriptions per image. The overall ranking is obtained by averaging the rankings from CHAIR \citep{chair} and POPE (MSCOCO subset) \citep{pope}.\footnote{CHAIR and POPE are the two popular representative metrics for free-form text generation and binary classification hallucination benchmarks respectively \citep{amber}.} In addition to SelfCheckGPT, UniHD\citep{chen2024unified} is used as a stronger baseline.

For evaluation, we take a subset of 30 image descriptions generated by each target model (a total of 270 passages with 3237 facts) and annotate each description with a binary label of either \textit{hallucinatory} or \textit{factual}. The Cohen's $\kappa$ between the two annotators is 0.632, indicating substantial agreement. The models are ranked by the average percentage of factual errors produced by each target model, and hallucination ranking performance is measured at the \emph{system-level} using SRC, denoted System($\rho$) and at the \emph{image-level} using PCC, denoted as Image($r$).

\begin{table}[h]
    \centering
    \begin{tabular}{lcccc}
    \toprule
    \multirow{2}{*}{Metrics}     & \multicolumn{3}{c}{System($\rho$) (\%)} & Image($r$) (\%) \\
                                 & Overall & CHAIR & Human & Human \\ 
    \midrule
    UniHD \citep{chen2024unified} & 42.02 & 36.98 & 33.33 & 36.70 \\
    SelfCheckGPT \citep{manakul2023selfcheckgpt} & 43.70 & 23.10 & -10.00 & 20.93 \\
    \midrule
    CrossCheck-implicit & \textbf{50.42} & \textbf{64.71} & \textbf{98.33} & \underline{48.72} \\
    CrossCheck-explicit & 42.86 & 43.70 & \underline{75.00} & 35.16 \\
    CrossCheck-implicit weighted & \textbf{50.42} & \textbf{64.71} & \textbf{98.33} & \textbf{52.83} \\
    CrossCheck-explicit weighted & \underline{47.06} & \underline{46.22} & 73.33 & 36.98 \\
    \bottomrule
    \end{tabular}
    \vspace{0.2cm}
    \caption{System-level correlation measured by System($\rho$) and Image-level correlation measured by Image($r$) for various hallucination evaluation methods on the MHaluBench dataset. System-level correlation is measured against the overall ranking, rankings from CHAIR scores and human annotation.}
    \vspace{-0.3cm}
    \label{tab:image_results}
\end{table}

\textbf{Hallucination Ranking Results}: Similar to before, Table \ref{tab:image_results} presents the system-level and image-level correlations against overall rankings and rankings derived from human annotations. Both variants of CrossCheckGPT outperform SelfCheckGPT and UniHD, with CrossCheck-implicit weighted performing best out of all methods, achieving a 98.33\% correlation with the rankings from human annotations. Equivalent statistical significance analysis and scatter plots are shown in Table \ref{tab:significance} and Fig. \ref{fig:imagescatter} in the Appendix \ref{sec:imageapp}, respectively. 

\subsection{Video-to-text Experiments}
\label{section:audiovisual}

Next, we apply CrossCheckGPT to AVHalluBench to investigate hallucination ranking in audio-visual LLMs. We consider 7 models that can handle video inputs and 6 models that can handle audio inputs. Three models, FAVOR \citep{avsalmonn}, Video-LLaMA \citep{videollama}, and Gemini 1.5 Pro \cite{gemini}, are in the intersection of the two sets, and can handle audio-visual inputs. When ranking hallucinations for visual description, we consider audio-visual LLMs with \textit{visual-only} inputs and \textit{audio-visual} inputs as separate systems, and hence, there are $7\!+\!3\!=\!10$ target models for ranking. We conduct a similar ranking scheme for audio descriptions, where there are $6\!+\!3\!=\!9$ target models. 
All the target models are also used as evidence models in CrossCheck-explicit,\footnote{Gemini 1.5 Pro is not used for CrossCheck-implicit due to the number of request limitations.} and each model generates ten evidence passages. When using audio-visual LLMs as evidence models, audio-visual inputs are given to obtain the visual or audio descriptions as evidence. 
As only 5 target models can handle speech inputs, we further make a dedicated ranking only for these models with prompts explicitly asking for speech description. 


\begin{table}[h]
    \centering
    \begin{tabular}{lcc|cc}
    \toprule
    \multirow{2}{*}{Metrics} & \multicolumn{2}{c}{Visual Description (\%)} & \multicolumn{2}{c}{Audio Description (\%)} \\
     & System($\rho$) & Video($r$) & System($\rho$) & Video($r$) (w. speech)\\
    \midrule
    SelfCheckGPT     & 86.67 & 65.77 & 60.00 & 51.13 (44.55) \\
    CrossCheck-implicit weighted & 54.29 & 30.73 & 40.00 & 2.15 (16.20) \\
    CrossCheck-explicit weighted & \textbf{89.09} & \textbf{78.58} & \textbf{71.67} & \textbf{68.10} (\textbf{47.60}) \\
    \bottomrule
    \end{tabular}
    \vspace{0.2cm}
    \caption{System-level and video-level correlations of SelfCheckGPT and CrossCheckGPT against RefCheck using manual descriptions in AVHalluBench. Weighted version of CrossCheckGPT is used with $C=0.1$. Ranking correlations for systems that handle speech are in brackets.}
    \vspace{-0.3cm}
    \label{tab:correlation_video}
\end{table}

\textbf{Hallucination Ranking Results}: First, system-level and video-level correlations are shown in Table~\ref{tab:correlation_video}, measured by System($\rho$) and Video($r$). CrossCheck-explicit correlates with RefCheck best, with an 89.09\% System($\rho$) for the visual description. Similar to the text-to-text results, we observe that CrossCheck-explicit performs better than CrossCheck-implicit. For both text-to-text and video-to-text experiments, this is likely due to the high diversity in the evidence passages as indicated by high raw SelfCheckGPT scores, which we discuss further in Section \ref{ssec:explicit_v_implicit}.





\textbf{Impact of Audio-Visual Inputs}: As supporting information from another modality is expected to reduce hallucination, this section investigates whether audio-visual inputs reduce the raw hallucination scores compared to the scores when a single modality is used. Table \ref{tab:audiovisual} presents the average raw hallucination scores (rather than correlations), for three MLLMs that can take audio-visual inputs. 
\begin{table}[!ht]
    \centering
    \begin{tabular}{lccccc}
    \toprule
    \multirow{2}{*}{Model} & \multirow{2}{*}{Input modality} & \multicolumn{2}{c}{Visual Description (\%)} & \multicolumn{2}{c}{Audio Description (\%)} \\
     &  & $\mathcal{S}_\text{selfcheck}$ $\downarrow$ & $\mathcal{C}_\text{explicit}$ $\downarrow$ & $\mathcal{S}_\text{selfcheck}$ $\downarrow$ & $\mathcal{C}_\text{explicit}$ $\downarrow$ \\
    \midrule
    \multirow{3}{*}{FAVOR~\cite{avsalmonn}} & Visual & 60.67 & 53.85 & --- & --- \\
     & Audio & --- & --- & {49.62} & 66.69 \\
     & Audio-Visual & 56.42 & {49.60} & {33.25} & {35.20} \\
    \midrule
    \multirow{3}{*}{Video-LLaMA~\cite{videollama}}  & Visual & {41.14} & 52.02 & ---&--- \\
     & Audio & ---&--- & 56.42 & 68.05 \\
     & Audio-Visual & {47.73} & {49.13} & 70.23 & {41.25} \\
    \midrule
    \multirow{3}{*}{Gemini 1.5 Pro~\cite{gemini}} & Visual & {19.87} & {31.74} & --- & --- \\
    & Audio  & --- & --- & \textbf{25.82} & {34.66} \\
    & Audio-Visual & \textbf{12.77} & \textbf{23.27} & 48.51 & \textbf{28.79} \\
    \bottomrule
    \end{tabular}
    \vspace{0.2cm}
    \caption{SelfCheckGPT scores ($\mathcal{S}_\text{selfcheck}$) and weighted CrossCheck-explicit scores ($\mathcal{C}_\text{explicit}$) on AVHalluBench for audio-visual LLMs. Calibration temperature $T=0.1$ is used here.}
    \vspace{-0.3cm}
    \label{tab:audiovisual}
\end{table}

When considering the CrossCheckGPT scores, we observe that having audio-visual inputs reduces hallucination rates, as measured by the raw CrossCheckGPT scores, as expected. While Gemini 1.5 Pro achieved the best scores, it can be more susceptible to hallucination when silent videos are used as inputs as it often fabricates its audio descriptions. Moreover, except for Gemini 1.5 Pro, when audio-visual inputs are used the reduction in hallucination scores is larger for audio description tasks than for visual description tasks. This likely occurs as for audio description tasks, visual information often provides useful information on the source of the sound, which can significantly reduce the uncertainty of the sound. For visual description tasks, while particular audio cues (especially from speech) can provide useful information, misleading or unrelated sounds may cause additional hallucinations. For example, in Fig \ref{fig:example3} where there is a self-playing piano, audio inputs can mislead a model to believe that the piano is played by an individual. Further examples are presented in Appendix \ref{sec:AVHalluBench} with the raw hallucination scores for audio and visual-only inputs shown in Tables \ref{tab:video_results} and \ref{tab:audio_results} in Appendix.


\subsection{CrossCheck-explicit vs. CrossCheck-implicit}
\label{ssec:explicit_v_implicit}
While CrossCheck-implicit is more sample-efficient than CrossCheck-explicit and only requires generating the error analysis once, the performance of CrossCheck-implicit can be highly dependent on the task. For the text-to-text and video-to-text experiments, CrossCheck-implicit performs worse than CrossCheck-explicit, as opposed to the findings in the image-to-text experiments. We hypothesize that for challenging and open-ended tasks, CrossCheck-explicit is preferred as it can better cover the output space by disentangling the evidence generation and verification tasks, yielding more calibrated uncertainty measures. However, in other circumstances, CrossCheck-implicit may help the model focus on specific aspects of the input and yield more accurate rankings. For challenging and open-ended tasks with diverse outputs, the raw SelfCheckGPT scores are expected to be high and therefore can be used as a proxy to determine which consistency measure to select. For example, the average SelfCheckGPT score across models is 40.63\% for text-to-text, which is much higher than 17.16\% for image-to-text. We recommend using CrossCheck-explicit when the SelfCheckGPT scores are high, and CrossCheck-implicit when they are sufficiently low, which is demonstrated to be a reasonable rule, illustrated by the results in Appendix Table \ref{tab:recommendation}.

\subsection{Ablation Studies}


\textbf{Self-Bias}: LLMs are known to have self-preferential bias \cite{brown1986evaluations, zheng2024judging} and may prefer outputs from similar models. Therefore LLMs using the same base model may provide inflated CrossCheckGPT scores. The results in Table \ref{tab:selfbias} show that self-bias is an issue, and for example, when only using Llama-2-based evidence models, the outputs from Vicuna get a lower hallucination score whereas when only using Mistral-based evidence models, Mistral has the lowest hallucination score, resulting in contradictory conclusions. This bias can be mitigated by adopting a wide range of evidence models, which is adopted in CrossCheckGPT scores, hence achieving more reliable evaluation with strong correlations.

\begin{table}[h]
    \centering
    \begin{tabular}{lcccc}
    \toprule
    Evidence Models & System($\rho$) & Document($r$) & Vicuna $\mathcal{C}_\text{explicit}$ & Mistral $\mathcal{C}_\text{explicit}$ \\
    \midrule
    Llama-2-based models only & 55.49\% & 81.10\% & 42.94\% & 45.68\% \\
    Mistral-based models only & 81.71\% & 81.06\% & 44.98\% & 41.81\% \\
    All models & {82.32\%} & {82.28\%} & 44.82\% & 44.93\% \\
    \bottomrule
    \end{tabular}
    \vspace{0.2cm}
    \caption{The mitigation of self-bias in CrossCheckGPT scores and its influence measured by document-level correlations and CrossCheck-explicit scores of Vicuna and Mistral on WikiBio. There are 4 Llama-2-based models and 4 Mistral-based models in the set of evidence models.}
    \vspace{-0.5cm}
    \label{tab:selfbias}
\end{table}

\begin{figure}[h]
    \centering
    \includegraphics[scale=0.48]{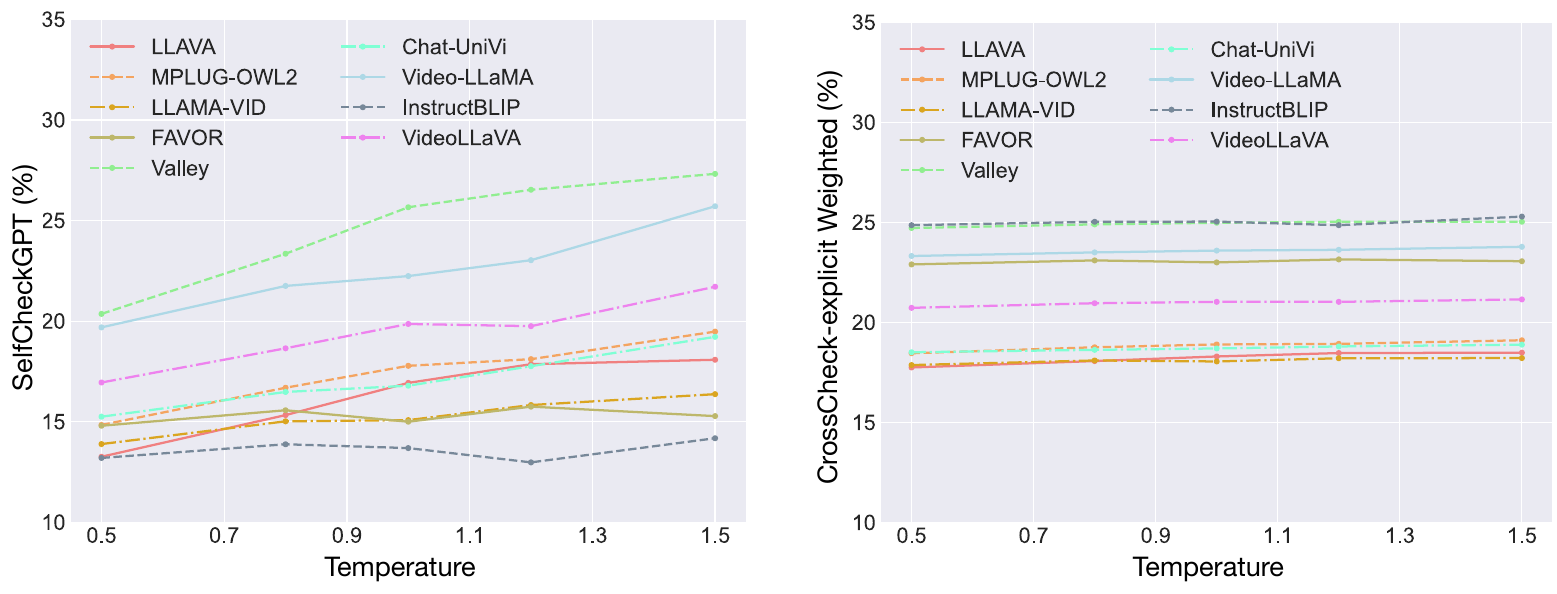}
    \vspace{-0.2cm}
    \caption{Variation of SelfCheckGPT scores (Left) and the weighted CrossCheck-explicit scores (Right) against the varying temperature during description generation.}
    \vspace{-0.2cm}
    \label{fig:crosscompare}
\end{figure}
\textbf{Robustness to Manipulation}: To investigate whether a ranking method can be easily manipulated, we examine the influence of the generation temperature (which can be selected for any model). The results in Fig. \ref{fig:crosscompare} show that by increasing the temperature of the target model from 0.5 to 1.5, SelfCheckGPT scores increase by as much as 35\%, drastically influencing the rankings. In contrast, CrossCheckGPT provides more stable rankings for all generation temperatures. Results are demonstrated for MHaluBench, but similar trends are observed for WikiBio as well.

\section{Conclusions}

This paper proposes CrossCheckGPT, a universal hallucination ranking method for multimodal large language models. We evaluated two variants of CrossCheckGPT on text-to-text, image-to-text and video-to-text tasks, demonstrating that it consistently outperforms all baseline methods, achieving 98\% and 89\% system-level correlation against humans on MHaluBench and AVHalluBench respectively. We also introduce AVHalluBench, the first resource to study audio-visual hallucination issues in video understanding.



\section*{Acknowledgments}
This work is supported by Cambridge University
Press \& Assessment (CUP\&A), a department of The Chancellor, Masters, and Scholars of the University of Cambridge.

{
\small
\bibliographystyle{abbrvnat}
\bibliography{references,acl_anthology}
}

\newpage
\appendix
\section{Experimental Setup Details}
\label{sec:models}
We list the models involved in this paper in Table \ref{tab:models}, and text-to-text metrics in Table \ref{tab:benchmark}.
\begin{table}[h]
    \footnotesize
    \centering
    \begin{tabular}{lcccc}
    \toprule
     Target LLMs & Modality & Evidence Models & Evidence Models & License  \\
     & & (explicit) & (explicit) \\
    \midrule
    Llama-2-7B \cite{manakul2023selfcheckgpt} & Text & \ding{51} & \ding{51} & llama2 \\
    Llama-2-7B-Chat \cite{manakul2023selfcheckgpt} & Text & \ding{51} & \ding{51} & llama2 \\
    Mistral-7B-Instruct-v0.1 \cite{jiang2023mistral} & Text & \ding{55} & \ding{55} & Apache-2.0 \\
    Mistral-7B-Instruct-v0.2 \cite{jiang2023mistral} & Text & \ding{51} & \ding{51} & Apache-2.0 \\
    Vicuna-v1.5-7B\cite{vicuna} & Text & \ding{51} & \ding{51} & llama2 \\
    Falcon-7B\cite{falcon} & Text & \ding{55} & \ding{55} & Apache-2.0 \\
    Starling-7B-alpha\cite{starling2023} & Text & \ding{51} & \ding{51} & Apache-2.0 \\
    StableBeluga-7B\cite{StableBeluga} & Text & \ding{51} & \ding{51} & llama2 \\
    Zephyr-7b-beta\cite{zephyr} & Text & \ding{51} & \ding{51} & MIT \\
    Mistral-7B-OpenOrca\cite{openorca} & Text & \ding{51} & \ding{51} & Apache-2.0 \\
    GPT-4 \cite{gpt4} & Text & \ding{55} & \ding{55} & N/A \\
    \midrule
    LLaVA-v1.5 \cite{llava} & Vision & \ding{51} & \ding{51} & llama2 \\
    InstructBLIP (vicuna-7B) \cite{instructblip} & Vision & \ding{51} & \ding{55} & BSD 3-Clause \\
    mPLUG-Owl2 \cite{mplugowl2} & Vision & \ding{51} & \ding{51} & MIT \\
    Valley \cite{valley} & Vision & \ding{51} & \ding{51} & Apache-2.0 \\
    Video-LLaVA \cite{videollava} & Vision & \ding{51} & \ding{51} & Apache-2.0 \\
    Chat-Univi \cite{chatunivi} & Vision & \ding{51} & \ding{51} & Apache-2.0 \\
    LLaMA-VID \cite{llamavid} & Vision & \ding{51} & \ding{55} & Apache-2.0\\
    \midrule
    LTU \cite{ltu} & Audio & \ding{51} & \ding{51} & Apache-2.0 \\
    Qwen-Audio-Chat \cite{QwenAudio} & Audio & \ding{51} & \ding{51} & Tongyi Qianwen\\
    SALMONN \cite{salmonn} & Audio & \ding{51} & \ding{51} & Apache-2.0 \\
    \midrule
    Video-LLaMA \cite{videollama} & Audio-visual & \ding{51} & \ding{51} & BSD 3-Clause\\
    FAVOR \cite{avsalmonn} & Audio-visual & \ding{51} & \ding{51} & Apache-2.0 \\
    Gemini 1.5 Pro \cite{gemini} & Audio-visual & \ding{51} & \ding{55} & N/A \\
    \bottomrule
    \end{tabular}
    \vspace{0.2cm}
    \caption{Models and reference benchmarks for validating CrossCheckGPT.}
    \label{tab:models}
\end{table}

\begin{table}[h]
    \footnotesize
    \centering
    \begin{tabular}{p{5cm}p{9cm}}
    \toprule
     Reference Benchmarks (Metrics) & Description \\
    \midrule
    TriviaQA \cite{triviaqa} (Acc) & A realistic text-based question-answering dataset containing documents collected from Wikipedia and the web. \\
    \midrule
    TruthfulQA MC1 \cite{truthfulqa} (Acc) & \multirow{2}{9cm}{A benchmark to measure whether a language model is truthful in generating answers to questions, spanning 38 categories.}\\
    TruthfulQA MC2 \cite{truthfulqa} (Acc) & \\
    \midrule
    XSum \cite{xsum} (FactKB \cite{factkb}) & The factual accuracy of summarization models by verifying the presence of knowledge base facts in generated summaries.\\
    \midrule
    CNN-DM \cite{cnndm} (BERTP) & The CNN-DailyMail dataset is a collection of news articles and accompanying summaries measured by BERTScore-Precision. \\
    \midrule
    MemoTrap \cite{memotrap} (Acc) & Assessing whether LLMs fall into memorization traps which occur when LLMs memorize specific examples in training.\\
    \midrule
    FaithDial \cite{faithdial} (Acc) & A benchmark for hallucination-free dialogues by editing hallucinated responses in Wizard of Wikipedia (WoW) \cite{wizard}\\
    \midrule
    HaluEval-QA \cite{halueval} (Acc) & \multirow{2}{9cm}{A large collection of generated and human-annotated hallucinated samples for evaluating the performance of LLMs in recognizing hallucination. It contains the QA, summarization and dialogue tasks.} \\
    HaluEval-summarization \cite{halueval} (Acc) & \\
    HaluEval-Dialogue \cite{halueval} (Acc) & \\
    \bottomrule
    \end{tabular}
    \vspace{0.2cm}
    \caption{Dataset, models and reference benchmarks for validating CrossCheckGPT. Acc stands for accuracy.}
    \label{tab:benchmark}
\end{table}

\section{Exact Prompts}
We provide the exact prompts we used in our experiments in Table \ref{tab:prompt} for various tasks.
\begin{table}[h]
    \centering
    \begin{tabular}{lp{7cm}}
    \toprule
    Task     & Prompt \\
    \midrule
    Text-to-text generation & Generate a passage about $<$name$>$.\\
    Image-to-text description & Describe the image in one paragraph. \\
    Visual description for video & Describe the video in one paragraph.\\
    Audio description for video & Describe the audio in one paragraph. \\
    Prompt for speech content & What does the man/woman say in the video?\\
    \midrule
    LLM Judgment for CrossCheck-explicit  & Context: $<$evidence\_passage$>$$\backslash$n$\backslash$nSentence: $<$sentence$>$ $\backslash$n$\backslash$nIs the sentence supported by the context above? Answer Yes or No.$\backslash$n$\backslash$nAnswer:  \\
    CrossCheck-implicit factual errors & You are given the following sentence about $<$name/image/video$>$ that might be inaccurate:$\backslash$n$<$sentence$>$$\backslash$n List possible inaccurate information in this sentence.\\
    LLM Judgment for CrossCheck-implicit  & You are given the following sentence about $<$name/image/video$>$:$\backslash$n$<$sentence$>$$\backslash$nThe following is an analysis of possible inaccuracies in this sentence:$\backslash$n$<$list\_of\_possible\_errors$>$$\backslash$nBased on the analysis, determine if the sentence contains any inaccurate information. Answer Yes or No.$\backslash$n$\backslash$nAnswer: \\
    \bottomrule
    \end{tabular}
    \vspace{0.2cm}
    \caption{Exact prompt used for different tasks.}
    \label{tab:prompt}
\end{table}

\section{CrossCheckGPT as a Hallucination Detection Method}
CrossCheckGPT can be used as a Hallucination detection method, which performs better than the best output-probability-based method reported in SelfCheckGPT\cite{manakul2023selfcheckgpt}.
\begin{table}[!h]
    \centering
    \begin{tabular}{lcccc}
    \toprule
    Evidence Model & Non-Factual & Non-Factual* & Factual & Document ($r$) \\
    \midrule
    Llama 30B Max($\mathcal{H}$) \citep{manakul2023selfcheckgpt} &   80.92 & 37.32 & 37.90 & 35.57 \\
    \midrule
    Llama-2-7B-Chat & 85.84 & 57.22 & 54.41 & 56.25 \\
    Vicuna-v1.5-7B & 83.13 & 53.38 & 51.13 & 54.64 \\
    Mistral-7B-Instruct-v0.2 & \textbf{87.21} & \textbf{59.60} & \textbf{56.72} & \textbf{63.04} \\
    \bottomrule
    \end{tabular}
    \vspace{0.2cm}
    \caption{AUC-PR and document-level correlation against human annotation for detecting hallucinations in GPT-3 using individual evidence models on non-factual and factual statements in WikiBio~\cite{manakul2023selfcheckgpt}.}
    \label{tab:aucpr}
\end{table}

\section{Text-to-text Additional Results}
We provide the version of Table \ref{tab:weightcross} with all ten benchmark metrics in Table \ref{tab:weightcross2}. Moreover, we investigate the \textit{specific-task} hallucination ranking ability where the inputs to SelfCheckGPT and CrossCheckGPT are from a specific task (rather than text generation). We conduct task-specific experiments using the inputs from TruthfulQA MC1 and HaluEval QA containing multiple-choice and yes-no questions respectively. The results in Table~\ref{tab:taskspecific} show high system-level correlations and moderate document-level correlations, indicating that CrossCheckGPT can operate as a task-specific metric without requiring any ground truth.
\begin{table}[h]
    \small
    \centering
    \begin{tabular}{p{5cm}ccc}
    \toprule
    \multirow{2}{*}{Metrics}     & \multirow{2}{*}{System($\rho$)} & \multicolumn{2}{c}{Document ($r$)} \\
                                 & &w/o GPT4 & with GPT4 \\ 
    \midrule

    TriviaQA \cite{triviaqa}              & 23.33  & - & - \\
    TruthfulQA MC1 \cite{truthfulqa}      & 52.94  & - & - \\
    TruthfulQA MC2 \cite{truthfulqa}      & 57.14  & - & - \\
    XSum \cite{xsum}              & -70.00 & - & - \\
    CNNDM \cite{cnndm}            & 38.33  & - & - \\
    MemoTrap \cite{memotrap}      & 10.88  & - & - \\
    FaithDial \cite{faithdial}    & -8.33  & - & - \\
    HaluEval-QA \cite{halueval}   & -18.33 & - & - \\
    HaluEval-Summarization \cite{halueval} & 48.33 & - & - \\
    HaluEval-Dialogue \cite{halueval} & 46.03 & - & - \\
    SelfCheckGPT \cite{manakul2023selfcheckgpt} & 66.46 & 74.06 & 76.08 \\
    \midrule
    CrossCheck-explicit & \underline{77.44} & \textbf{82.28} & \underline{77.23} \\
    CrossCheck-implicit & 56.71 & 18.33 & 17.29 \\
    CrossCheck-explicit weighted & \textbf{82.32} & \underline{81.78} & \textbf{82.18} \\
    CrossCheck-explicit weighted & 56.81 & 20.21 & 19.16 \\
    \bottomrule
    \end{tabular}
    \vspace{0.2cm}
    \caption{Full version of Table \ref{tab:weightcross} including all other metrics. General hallucination evaluation where the task for SelfCheckGPT/CrossCheckGPT is open-ended text generation on WikiBio. System-level correlation, System($\rho$), is measured against the overall ranking in the leaderboard, and document-level correlation, Document($r$), is measured against RefCheck. With GPT-4 refers to including GPT-4 as the target LLM.}
    \label{tab:weightcross2}
\end{table}

\begin{table}[!ht]
    \centering
    \begin{tabular}{lcccc}
    \toprule
    \multirow{2}{*}{Metrics}     & \multicolumn{2}{c}{System($\rho$)} & \multicolumn{2}{c}{Document ($r$)} \\
                                 & TruthfulQA MC1 & HaluEval QA & TruthfulQA MC1 & HaluEval QA \\
    \midrule
    SelfCheckGPT     & \textbf{76.19} &  30.95 & 30.87 & 6.76 \\
    CrossCheckGPT     & \textbf{76.19} & \textbf{88.10} & \textbf{33.68}  & \textbf{22.00} \\
    \bottomrule
    \end{tabular}
    \vspace{0.2cm}
    \caption{Task-specific hallucination evaluation where the task of SelfCheckGPT/CrossCheckGPT is, in this example, either TruthfulQA MC1 or HaluEval QA. Note that rankings are performed on 8 target models that are instruction-tuned as these tasks are QA-based and require some instruction-following ability.}
    \label{tab:taskspecific}
\end{table}


\begin{figure}[!ht]
    \centering
    \includegraphics[scale=0.5]{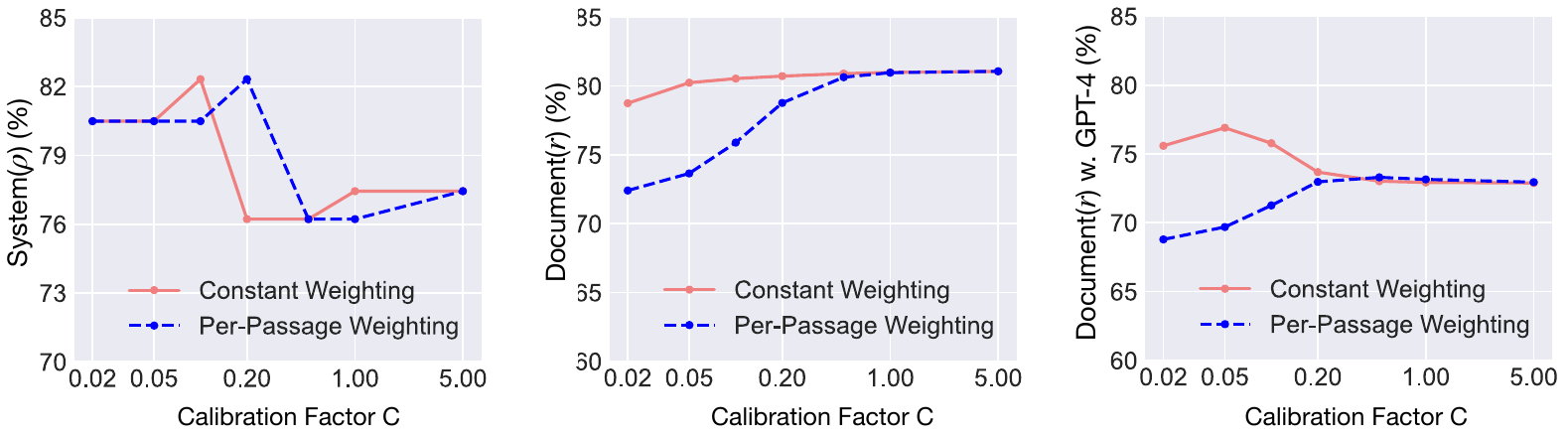}
    \caption{The variation of System($\rho$) and Document($r$) against calibration temperature $T$ in Eqn. \eqref{eq:weighting} for weighted CrossCheck-explicit. Constant weighting refers to applying the same weight for all documents, while per-passage weighting refers to the use of passage-specific weighting derived from SelfCheckGPT scores of each passage.}
    \label{fig:textcalibration}
\end{figure}
 We first show the \textbf{variation of system and document-level correlation against varying calibration temperatures} for CrossCheck-explicit weighted in Fig. \ref{fig:textcalibration} using WikiBio data. A comparison between using per-query weights and using the same weights for the entire task is also provided. As a result, $C=0.1$ is chosen as it achieves the best system-level correlation. Besides, the same weighting across the whole task is used at $C=0.1$ as the large variance among weights of different queries introduces more noise in scoring and hence hinders the correlation.

\section{System-level Correlations between Individual Text-based Hallucination Benchmarks}

We provide the system-level correlations between individual text-based hallucination benchmarks to show that they capture different aspects and do not correlate well with each other in Table \ref{tab:metricssystem}.

\begin{table}[h]
    \centering
    \scriptsize
    \begin{tabular}{lcccccccccc}
    \toprule
     & TriviaQA & TruthfulQA	 & Xsum & CNN-DM & MemoTrap & FaithDial & HaluQA & HaluSumm & HaluDial \\
    \midrule
    TriviaQA \cite{triviaqa}   &  1.00 & 0.20  & -0.72 & 0.15 & 0.07 & 0.13 & 0.27 & 0.40 & 0.50\\
    TruthfulQA \cite{truthfulqa} & 0.20 & 1.00 & -0.10 & 0.38 & 0.27 & 0.05 & -0.50 & 0.37 & 0.63 \\
    Xsum \cite{xsum} & -0.72 & -0.10  & 1.00 & -0.03 & -0.40 & 0.12 & -0.57 & -0.63 & -0.68 \\
    CNN-DM \cite{cnndm} & 0.15 & 0.38  & -0.03 & 1.00 & 0.28 & -0.05 & -0.05 & 0.33 & 0.37 \\
    MemoTrap \cite{memotrap}  & 0.07 & 0.27  & -0.40 & 0.28 & 1.00 & -0.05 & -0.08 & 0.48 & 0.17 \\
    FaithDial \cite{faithdial} & 0.13 & 0.05  & 0.12 & -0.05 & -0.05 & 1.00 & -0.03 & -0.22 & -0.13 \\
    HaluQA \cite{halueval} & 0.27 & -0.50 & -0.57 & -0.05 & -0.08 & -0.03 & 1.00 & 0.30 & 0.20 \\
    HaluSumm \cite{halueval} & 0.40 & 0.37 & -0.63 & 0.33 & 0.48 & -0.22 & 0.30 & 1.00 & 0.67 \\
    HaluDial \cite{halueval} & 0.50 & 0.63 & -0.68 & 0.37 & 0.17 & -0.13 & 0.20 & 0.67 & 1.00 \\
    \bottomrule
    \end{tabular}
    \vspace{0.2cm}
    \caption{System-level correlation ($\rho$) between each pair of the 9 selected benchmarks metrics.}
    \label{tab:metricssystem}
\end{table}

\section{Scatter Plots and Statistical Significance for Image-to-text}
\label{sec:imageapp}

\begin{figure}[h]
    \centering
    \includegraphics[scale=0.5]{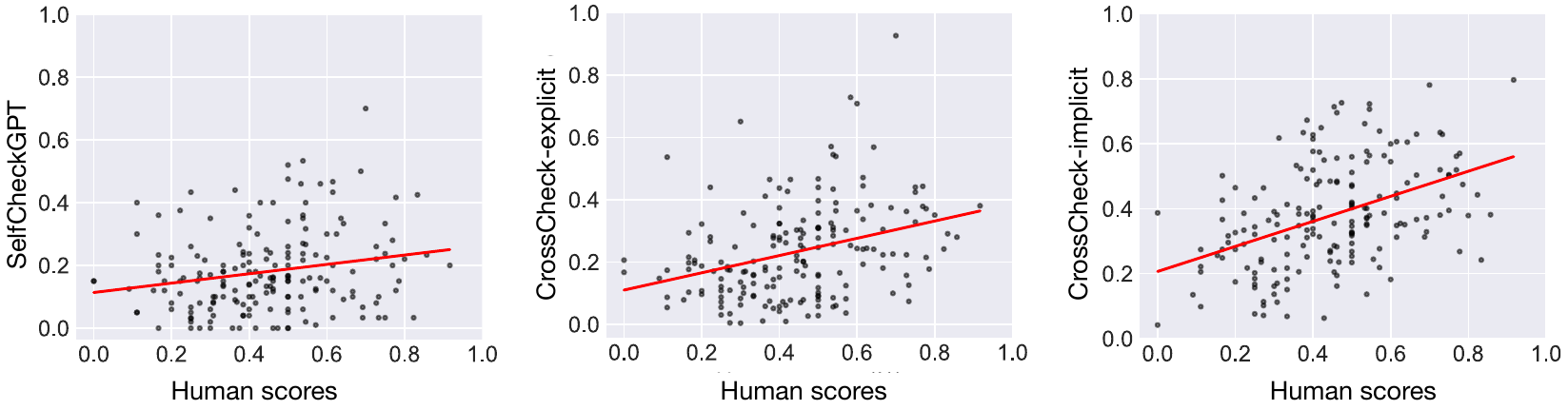}
    \caption{Scatter plot of SelfCheckGPT, CrossCheck-explicit and CrossCheck-implicit scores against human annotation for image-to-text tasks.}
    \label{fig:imagescatter}
\end{figure}

The scatter plot, similar to text-to-text ones in Fig. \ref{fig:textscatter}, is shown in Fig. \ref{fig:imagescatter}.

\begin{table}[h]
    \centering
    \begin{tabular}{lcccc}
    \toprule
    Methods     & Success rate (p-value) \\
    \midrule
     CrossCheck-explicit & 65.5\% ($<$0.00001)  \\
     CrossCheck-implicit & 84.5\% ($<$0.00001) \\
     CrossCheck-explicit weighted & 67.0\% ($<$0.00001)\\
     CrossCheck-implicit weighted & \textbf{88.0\%} ($<$0.00001)  \\
    \bottomrule
    \end{tabular}
    \vspace{0.2cm}
    \caption{Success rate and statistical significance of CrossCheckGPT approaches measured via sign-test on independent subsets of images.}
    \label{tab:significance}
\end{table}

Additionally, we report the statistical significance of CrossCheckGPT being better than SelfCheckGPT on MHaluBench by performing the sign test at the image level.

\section{Statistics of AVHalluBench}
We provide detailed statistics about AVHallubench in Table \ref{tab:AVHalluBench1}, including the number of videos, average lengths of each subset, as well as various audio and visual elements involved.
\begin{table}[h]
    \small
    \centering
    \begin{tabular}{lccccc}
    \toprule
      Source Dataset & Num. of Videos & Avg. Length (sec.) & w/ Speech & w/ Music & w/ Visual Text\\
      \midrule
      NeXT-QA \cite{nextqa} & 32 (18\%) & 22.0 & 19 & 7 & 1 \\
      M3AV \cite{m3av} & 27 (16\%) & 11.3 & 27 & 0 & 27 \\
      How2 \cite{how2} & 27 (16\%) & 9.5 & 27 & 4 & 2 \\
      MUSIC-AVQA \cite{musicavqa} & 23 (13\%) & 29.0 & 0 & 23 & 0 \\
      VALOR32k \cite{valor} & 26 (15\%) & 8.7 & 11 & 7 & 8 \\
      FAVDBench \cite{favd} & 38 (22\%) & 8.0 & 8 & 15 & 13 \\
      \midrule
      Overall & 175 & 14.2 & 92 (52\%) & 56 (32\%) & 51 (29\%) \\
        \bottomrule
      \end{tabular}
      \vspace{0.2cm}
      \caption{Statistics of the AVHalluBench dataset with the percentage shown in brackets.}
      \vspace{-0.2cm}
      \label{tab:AVHalluBench1}
\end{table}

\section{Additional SelfCheckGPT and CrossCheckGPT Scores on AVHalluBench}
We provide the detailed SelfCheckGPT and CrossCheckGPT scores on AVHalluBench for all MLLMs that handle video or audio inputs in this paper in Table \ref{tab:video_results} for video descriptions and Table \ref{tab:audio_results} for audio descriptions.
\label{sec:AVHalluBench}
\begin{table}[h]
    \centering
    \begin{tabular}{lcccc}
    \toprule
    Models & SelfCheckGPT & CrossCheck-explicit & CrossCheck-implicit \\
    \midrule
    Valley \cite{valley} & 52.43 & 55.98 & 48.22 \\
    Video-LLaVA \citep{videollava} & {30.59} & {33.52} & \underline{40.57} \\
    Chat-Univi \cite{chatunivi} & \underline{29.40} & \underline{32.68} & 41.75 \\
    LLaMA-VID \cite{llamavid} & 38.61 & 39.14 & \textbf{40.48} \\
    Video-LLaMA \cite{videollama} & 41.14 & 52.02 & 48.80 \\
    FAVOR \cite{avsalmonn} & 60.67 & 53.85 & 50.49 \\
    Gemini 1.5 Pro & \textbf{19.87} & \textbf{31.74} & - \\
    \bottomrule
    \end{tabular}
    \vspace{0.2cm}
    \caption{SelfCheckGPT and CrossCheckGPT scores for 6 visual-LLMs that take video as inputs on AVHalluBench. Note that FAVOR, Video-LLaMA and Gemini 1.5 Pro are only given visual inputs. Gemini 1.5 Pro was not used for CrossCheck-implicit.}
    \label{tab:video_results}
\end{table}

\begin{table}[h]
    \centering
    \begin{tabular}{lcccccc}
    \toprule
    Models & \multicolumn{2}{c}{SelfCheck} & \multicolumn{2}{c}{CrossCheck-explicit} & \multicolumn{2}{c}{CrossCheck-implicit} \\
    & audio & w. speech & audio & w.speech & audio & w. speech\\
    \midrule
    LTU \cite{ltu} & \textbf{21.95} & - & \underline{37.44} & - & \underline{18.06} & - \\
    Qwen-Audio-Chat \citep{QwenAudio} & 36.57  & 37.08 & 43.66 & 43.41 & 20.21 & \underline{52.20} \\
    SALMONN \cite{salmonn} & 34.99 & \underline{34.80} & 42.21 & \underline{40.15} & 18.32 & \textbf{48.17} \\
    FAVOR \cite{avsalmonn} & 49.62 & 41.51 & 66.69 & 55.41 & 23.26 & 61.01 \\
    Video-LLaMA \cite{videollama} & 56.42 & - & 68.05 & - & \textbf{17.10} & - \\
    Gemini 1.5 Pro & \underline{25.82} & \textbf{27.38} & \textbf{34.66} & \textbf{36.52} & - & - \\
    \bottomrule
    \end{tabular}
    \vspace{0.2cm}
    \caption{SelfCheckGPT and CrossCheckGPT scores for 6 audio-LLMs on AVHalluBench. Note that FAVOR and Video-LLaMA are only given audio inputs. Gemini 1.5 Pro was not used for CrossCheck-implicit.}
    \label{tab:audio_results}
\end{table}

\section{CrossCheck-explicit vs. CrossCheck-implicit}
We present the average SelfCheckGPT scores on each task together with the system-level correlations in Table \ref{tab:recommendation} to support our recommendations on CrossCheck-explicit and CrossCheck-implicit.

\begin{table}[h]
    \centering
    \begin{tabular}{lc|cc}
    \toprule
      & &  \multicolumn{2}{c}{System($\rho$)} \\
    Tasks     & Ave. $\mathcal{S}_\text{selfcheck}$ & CrossCheck-explicit& CrossCheck-implicit  \\
    \midrule
    Text-to-text     & 40.63 & 77.44 & 56.71 \\
    Image-to-text & 17.16 & 42.86 & 50.42 \\
    Audio description & 39.91 & 71.67 & 40.00 \\
    Visual description & 42.14 & 89.09 & 54.29 \\
    \bottomrule
    \end{tabular}
    \vspace{0.2cm}
    \caption{SelfCheckGPT scores and system-level correlations using CrossCheck-explicit and CrossCheck-implicit on four tasks. The system-level correlation for audio and visual descriptions is measured against RefCheck, and that for text-to-text and image-to-text tasks are measured against overall ranking.}
    \label{tab:recommendation}
\end{table}

\section{Case Studies for Hallucination with Audio-Visual Inputs}

\begin{figure}[h]
    \centering
    \includegraphics[scale=0.46]{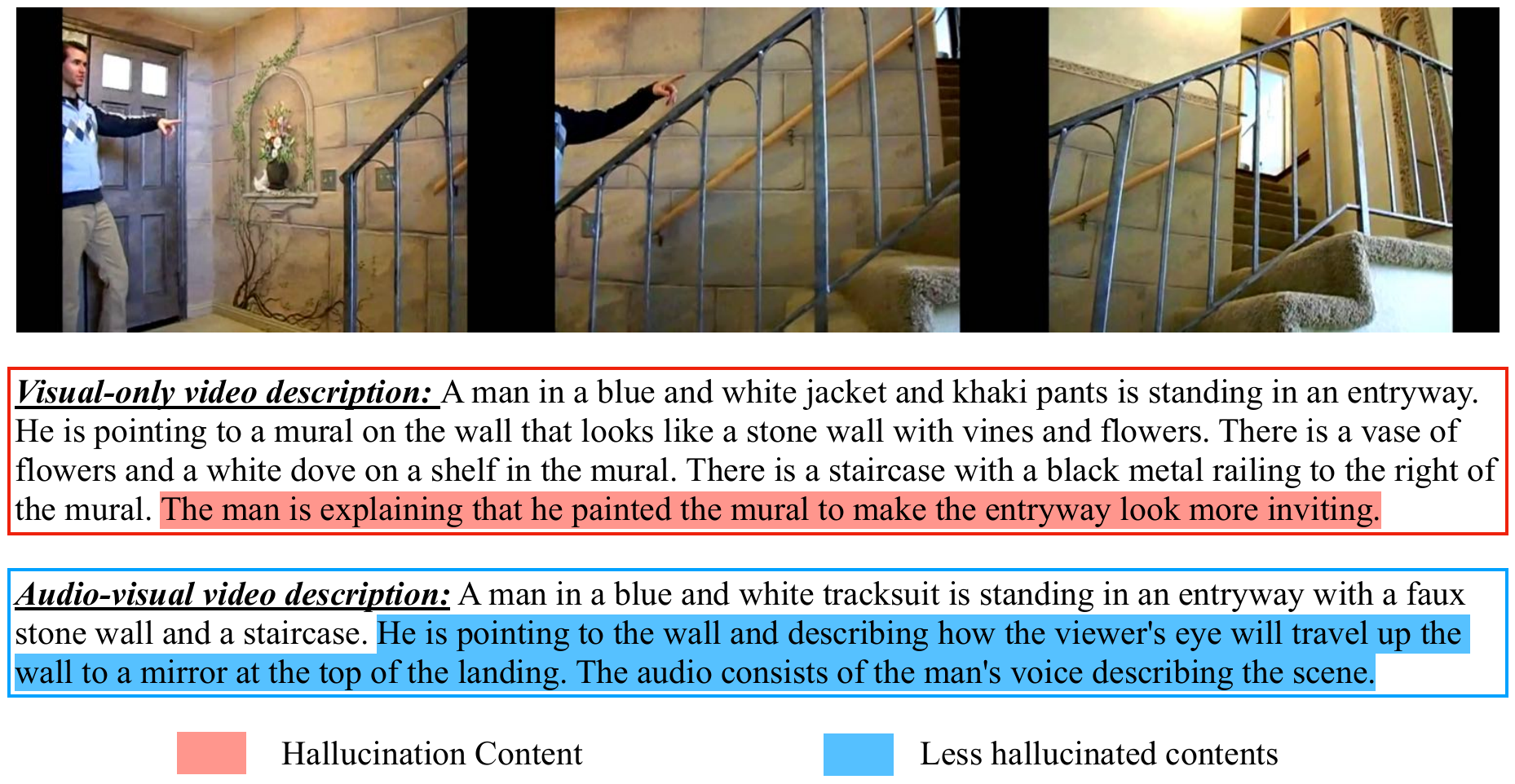}
    \caption{Example of audio-visual hallucination problem from Gemini 1.5 Pro. In this example, even when no audio is provided, the model still describes what the man is talking about, and having audio inputs greatly benefits the description by reducing the hallucination in describing the man's speech.}
    \label{fig:example2}
\end{figure}

\begin{figure}[h]
    \centering
    \includegraphics[scale=0.46]{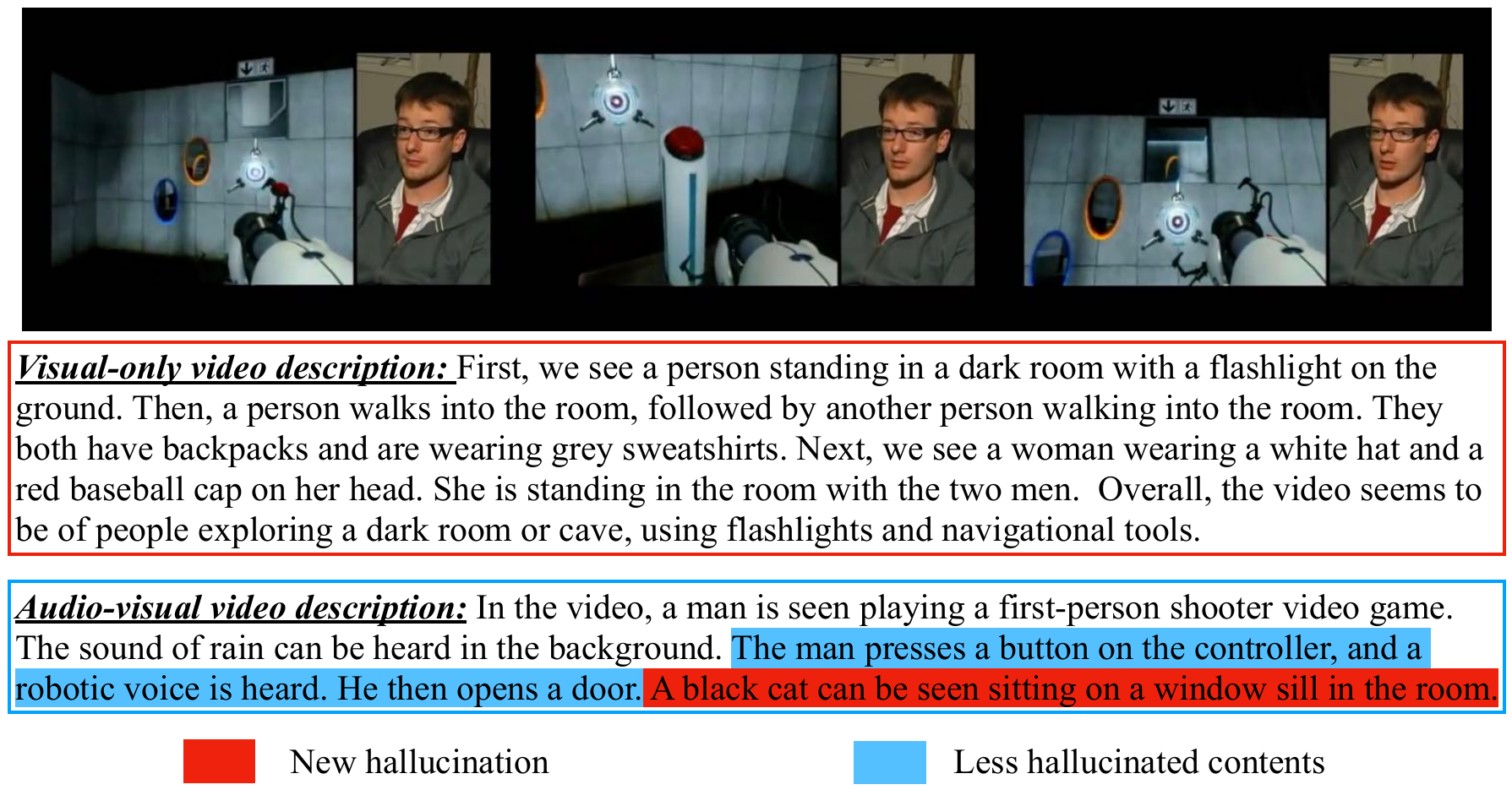}
    \caption{Example of audio-visual hallucination problem from FAVOR. In this example, the audio is the man explaining what he is doing in the game. The speech description reduces the hallucination of ``pressing the button'' and "opening a door" in the visual description with new but random hallucinations coming out.}
    \label{fig:example1}
\end{figure}

\begin{figure}[h]
    \centering
    \includegraphics[scale=0.46]{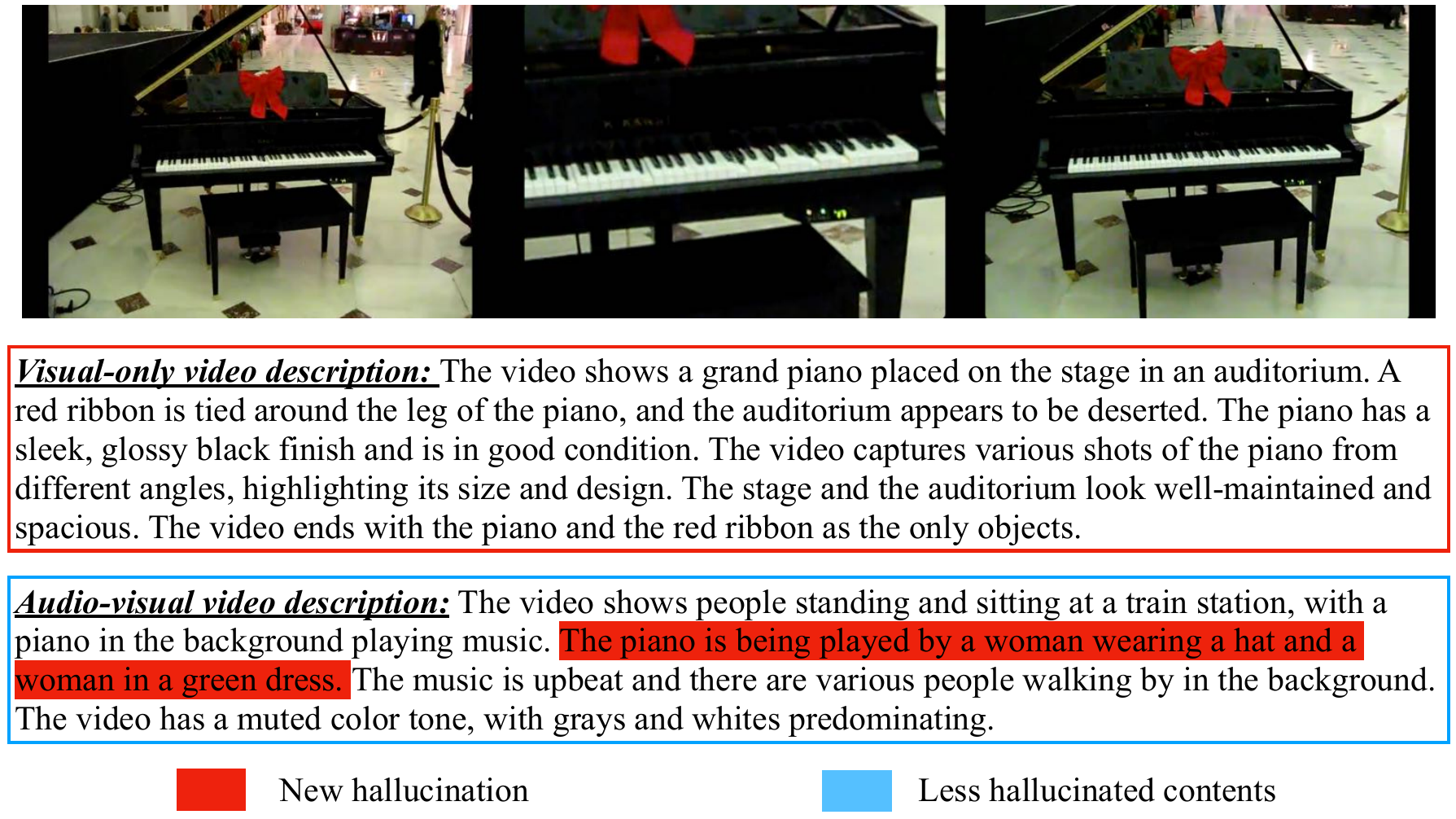}
    \caption{Example of audio-visual hallucination problem. In this example, the audio is the piano itself playing, which introduces additional hallucination to the visual description which describes it as ``played by a woman''.}
    \label{fig:example3}
\end{figure}

In addition to the piano example shown in Fig. \ref{fig:example3} that has been mentioned in the main text, we show here two additional examples in Fig. \ref{fig:example1} and Fig. \ref{fig:example2} where audio-visual inputs influence the hallucination compared to using audio or visual inputs alone. 

\newpage
\section{Limitations}
\label{sec:limit}
Our investigation is limited in the following aspects: First, hallucination is an expansive area and, as done in other studies, this paper only covers a reasonable subset of all possible domains. However, we plan to release a live hallucination leaderboard where we plan to benchmark the performance of further MLLMs over more benchmark metrics. Secondly, while the confidence-based weighting mechanism improves the performance of CrossCheckGPT, it does not take into account the similarities of different evidence models. Correlation between models, due to having similar training data or from starting at the same checkpoints, may result in evidence models making similar mistakes. This poses a future research direction, in raking model correlation into account for the weighting mechanism. Lastly, there is limited by the number of currently available audio-visual LLMs for evidence generation.


\section{Broader Impact}
\label{sec:broader}
Hallucinations in multimodal foundation models have become increasingly critical and challenging. Therefore, providing a general reference-free hallucination benchmarking approach is necessary and timely, enabling practitioners to have metrics for model trustworthiness. Therefore, CrossCheckGPT has the following positive broad impact:
\begin{itemize}
    \item CrossCheckGPT establishes a universal ranking system which helps identify more factual and faithful models to be selected in particular applications, reducing the dissemination of misinformation and increasing societal confidence in AI applications.
    \item CrossCheckGPT provides a reliable ranking that would aid regulatory bodies in enforcing compliance standards for multimodal foundation models, particularly in critical areas such as healthcare, finance, and public safety. 
    \item As a reference-free and versatile benchmarking method, CrossCheckGPT can drive developers to innovate and improve their multimodal foundation models.
\end{itemize}

However, our method by no means provides perfect hallucination scores and may inherit potential bias from the chosen evidence models. Therefore, practitioners should be independently educated and avoid overreliance on the rankings, as doing so may lead to complacency in critical thinking and reduced vigilance. From the model aspect, the approach in this paper does not give rise to any additional potential biases beyond the ones directly inherited from the pre-trained LLM checkpoints.

\section{Computing Resource}
\label{sec:computing}
Our experiments are performed on a single Nvidia A100 GPU for inference. The average inference time for each target model to get the CrossCheckGPT score is 20 hours. The total amount of time to run for all models in the text-to-text leaderboard is 200 hours, in the image-to-text leaderboard is 190 hours and in the AVHalluBench is 240 hours. The total GPU hours for running the full research is 2000. There is no training process involved in the research.

\section{Assets and License Explanation}
\label{sec:license}
Links to the following licenses that apply to the models used in the paper are provided (see Table \ref{tab:models}).
\begin{itemize}
    \item Llama2: \url{https://huggingface.co/meta-llama/Llama-2-7b-chat-hf/blob/main/LICENSE.txt}
    \item Apache-2.0: \url{https://www.apache.org/licenses/LICENSE-2.0}
    \item MIT License: \url{https://choosealicense.com/licenses/mit/}
    \item BSD 3-Clause License: \url{https://github.com/salesforce/LAVIS/blob/main/LICENSE.txt}
    \item Tongyi Qianwen: \url{https://github.com/QwenLM/Qwen-Audio/blob/main/LICENSE}
\end{itemize}

The following licenses are applied to the datasets used in our paper:
\begin{itemize}
    \item CC-BY-SA-3.0: Used by WikiBio hallucination data \cite{manakul2023selfcheckgpt}. License link: \url{https://spdx.org/licenses/CC-BY-SA-3.0}.
    \item MIT License: Used by MHaluBench (\url{https://huggingface.co/datasets/openkg/MHaluBench}). License link see above.
\end{itemize}

The following licenses are applied to the code and Python packages we use for our experiments:
\begin{itemize}
    \item Apache-2.0: Applies to Huggingface Transformers (\url{https://github.com/huggingface/transformers/blob/main/LICENSE}) and UniHD (\url{https://github.com/OpenKG-ORG/EasyDetect/blob/main/LICENSE}).
    \item MIT License: Applies to SelfCheckGPT (\url{https://github.com/potsawee/selfcheckgpt/blob/main/LICENSE}) and spaCy (\url{https://github.com/explosion/spaCy/blob/master/LICENSE}).
\end{itemize}



\end{document}